\documentclass[letterpaper]{article} 
\usepackage{aaai25}  
\usepackage{times}  
\usepackage{helvet}  
\usepackage{courier}  
\usepackage[hyphens]{url}  
\usepackage{graphicx} 
\usepackage{natbib}  
\usepackage{caption} 
\usepackage{array}
\usepackage{algorithm}
\usepackage{algorithmic}

\usepackage{newfloat}
\usepackage{listings}
\DeclareCaptionStyle{ruled}{labelfont=normalfont,labelsep=colon,strut=off} 
\floatstyle{ruled}
\newfloat{listing}{tb}{lst}{}
\floatname{listing}{Listing}
\nocopyright 

\title{Rethinking LLM Advancement: Compute-Dependent and Independent Paths to Progress}
\author{
    Jack Sanderson\textsuperscript{\rm 1},
    Teddy Foley\equalcontrib\textsuperscript{\rm 1},
    Spencer Guo\equalcontrib\textsuperscript{\rm 1},
    Anqi Qu\equalcontrib\textsuperscript{\rm 1},
    Henry Josephson\equalcontrib\textsuperscript{\rm 1}
}
\affiliations{
    \textsuperscript{\rm 1}Existential Risk Laboratory, University of Chicago\\

    \{jacksanderson, teddyfoley, scguo, anqiqu, henryj\}@uchicago.edu
}

\begin{document}

\maketitle

\begin{abstract}
Regulatory efforts to govern large language model (LLM) development have predominantly focused on restricting access to high-performance computational resources. This study evaluates the efficacy of such measures by examining whether LLM capabilities can advance through algorithmic innovation in compute-constrained environments. We propose a novel framework distinguishing compute-dependent innovations—which yield disproportionate benefits at high compute—from compute-independent innovations, which improve efficiency across compute scales. The impact is quantified using Compute-Equivalent Gain (CEG). Experimental validation with nanoGPT models confirms that compute-independent advancements yield significant performance gains (e.g., with combined CEG up to $3.5\times$) across the tested scales. In contrast, compute-dependent advancements were detrimental to performance at smaller experimental scales, but showed improved CEG (on par with the baseline) as model size increased, a trend consistent with their definition of yielding primary benefits at higher compute. Crucially, these findings indicate that restrictions on computational hardware, while potentially slowing LLM progress, are insufficient to prevent all capability gains driven by algorithmic advancements. We argue that effective AI oversight must therefore incorporate mechanisms for understanding, anticipating, and potentially guiding algorithmic research, moving beyond a singular focus on hardware. The proposed framework also serves as an analytical tool for forecasting AI progress.

\begin{links}
    \link{Code}{https://github.com/jcksanderson/llm-algorithmic-progress/}
\end{links}
\end{abstract}

\section{Introduction}

The rapid advancement of large language models (LLMs) has been driven by a few key factors, two of which are increases in computational resources and algorithmic improvements \cite{zhao_survey_2024, hoAlgorithmicProgressLanguage2024, sevilla2022computetrends, lohn_scaling_2023}. However, the relative interaction between these two drivers remain an open question with significant implications for AI progress, regulation, and forecasting. If new algorithmic advancements require commensurate computational resources, then restricting access to advanced hardware---through export controls or regulatory measures---could meaningfully slow AI development. Conversely, if algorithmic innovations can drive substantial progress even in a compute-limited environment, such restrictions may be far less effective than anticipated.

AI regulation has largely focused on hardware controls, particularly through export restrictions on cutting-edge chips \cite{fist2023preventing, gupta_whackachip_2024a}. Given that many recent LLM advancements have relied primarily on increasing scale through model size (parameter count), dataset size, and (more recently) inference time \cite{kaplan2020scalinglawsneurallanguage}, these restrictions prompt two fundamental questions. 
First, if computational power were frozen at current levels, could we still expect LLMs to continue improving? This speaks to the potential for progress even in a world where access to greater compute is fully restricted. Second, how do algorithmic or architectural advancements transfer between low compute and high compute? This question is more practical; rather than assuming perfect enforcement, it considers whether export controls could successfully prevent certain innovations from being leveraged at scale. 

Existing research \cite{epoch2024canaiscalingcontinuethrough2030, sevilla2022computetrends, hoAlgorithmicProgressLanguage2024, hernandezMeasuringAlgorithmicEfficiency, erdil_algorithmic_2023} has attempted to estimate the role of algorithmic progress through neural scaling laws. Performance (often measured by validation loss or perplexity) generally improves by investing greater computational resources (compute) in training, typically for the purpose of increasing both the number of model parameters and the size of the training dataset. However, past approaches suffer from two major limitations: scaling laws do not directly measure the contributions of specific algorithmic innovations, nor do they account for possible differences in the efficiency of algorithmic advancements at different compute scales. 

To address this, we introduce a new  framework that distinguishes between ``compute-dependent'' and ``compute-independent'' algorithmic advancements. Compute-dependent advancements' benefits emerge primarily with increases in compute, often becoming truly significant only at scales well beyond those of their initial conception, whereas compute-independent advancements enhance efficiency across all scales. 
To quantify the extent to which algorithmic advancements depend on compute, we estimate the compute-equivalent gain (CEG) \cite{davidsonAICapabilitiesCan2023} for each algorithm studied. Through a case study approach, we identify and analyze major algorithmic innovations from the past decade, classifying each as compute-independent/dependent and estimating their CEG. We then validate our framework by performing training experiments in two scales of nanoGPT, varying the use of several chosen algorithms and directly estimating their CEG.

Our findings suggest that while many of the most salient innovations in language modeling---such as the transformer architecture and mixture-of-experts models---have been compute-dependent, a large portion of LLM gains can be attributed to critical compute-independent advancements, such as rotary positional embeddings (RoPE), FlashAttention, and layer normalization. Importantly, the latter innovations improve model efficiency even in extremely resource-constrained settings, suggesting that further discovery of compute-independent algorithms could yield AI progress even in a compute-limited environment under hardware controls. Our experimental results further confirm that compute-independent advancements provide measurable performance gains in small-scale LLMs, while the benefits of compute-dependent advancements emerge as the compute scale increases.

By addressing these questions, this paper provides a clearer framework for policymakers, researchers, and industry leaders to navigate the complex interplay between computational resources and algorithmic innovation. Our analysis demonstrates the nuanced efficacy of hardware-centric AI governance: while such controls are unlikely to halt the progress propelled by compute-independent advancements (which can thrive even in resource-constrained environments), they may substantially influence the trajectory of compute-dependent breakthroughs, which often represent the most significant leaps in capability. Ultimately, this paper aims to equip stakeholders with a more precise understanding of these distinct modes of algorithmic progress, thereby informing more robust and adaptable AI governance strategies for the evolving technological landscape


\section{Methodology}

To systematically investigate the interplay between algorithmic progress and computational resources (particularly in the context of evaluating hardware controls), we employ a methodology centered on a novel analytical framework. Our approach is designed to disentangle gains from algorithmic innovation from those attributable to sheer increases in compute, and to understand how the benefits of different types of algorithmic advancements scale with computational capacity.

The core of our methodology is the distinction between compute-independent advancements---which enhance efficiency broadly across various computational scales---and compute-dependent advancements, whose most significant benefits are typically validated or fully realized at larger computational scales, even if their initial conceptualization occurred with less compute. To quantify and compare these different innovations, we utilize and adapt the concept of compute-equivalent gain (CEG), which measures algorithmic improvements in terms of equivalent computational expenditure.

The structure of our methodology is as follows:
\begin{enumerate}
  \item We first formally define our compute-independent/dependent framework and detail our approach to calculating CEG.
  \item We then apply this framework through case studies of impactful algorithmic advancements. This begins with an in-depth analysis of a contemporary example, DeepSeek-V3, followed by an examination of other major historical algorithms.
  \item Finally, we describe the setup for new empirical experiments conducted using nanoGPT models at two distinct scales. These experiments aim to directly measure CEG for selected compute-independent and compute-dependent algorithms, validating the predictions and utility of our framework.
\end{enumerate}

\subsection{Our Framework: Compute-Dependence and Compute-Equivalent Gain}\label{sec:ceg}
As previously discussed, increases in computational resources and hardware capabilities along with algorithmic breakthroughs have enabled the massive increases in LLM capabilities. Prior to the introduction of GPUs in deep learning in the early 2010s, compute levels used in AI increased more slowly than in the last decade or so \cite{sevilla2022computetrends}.

To understand the interaction between advances in computing resources and algorithmic innovations, we introduce our compute-dependent vs. compute-independent framework, as well as our methodology for estimating CEG. 

\begin{table*}[t]
\centering
\caption{Schematic for understanding compute-independent (I) and compute-dependent (D) algorithmic advancements.}
\begin{tabular}{|c|c|c|c|} 
\hline
 & \textbf{Status Quo Algorithm} & \textbf{New Algorithm (I)} & \textbf{New Algorithm (D)} \\ 
\hline
\textbf{Baseline Compute}  & Slow & Faster & Benefit Unrealized \\ 
\hline
\textbf{Increased Compute} & Slow & Faster & Very Fast\\
\hline
\end{tabular}
\label{tab:schema}
\end{table*}

\subsubsection{Compute-Independent vs. Compute-Dependent Algorithmic Advancements}
We classify an algorithmic advancement as compute-dependent or compute-independent by comparing the performance of the new algorithm (e.g., training procedure or architectural innovation) to its predecessor (i.e., status quo) algorithm at baseline and increased levels of compute. If an algorithm yields similar performance gains over a previous algorithm at both levels of compute, we interpret it as a compute-independent advance. 
Alternatively, if an algorithm provides small benefits (or worsens performance) at baseline compute, but gives large benefits with increased compute, we classify it as a compute-dependent advance. 

It is important to note two aspects of this dependency. First, this classification refers to where the most substantial benefits of an advancement are validated and fully realized, not necessarily where the advancement is initially conceived or discovered; indeed, compute-dependent innovations may often be first identified during smaller-scale experiments. Second, the emergence of these significant benefits in compute-dependent advancements is typically not tied to a single, universal compute threshold, but rather can occur progressively across a spectrum as computational scale increases. Thus, the distinction between `low' and `high' compute for observing these effects is often more of a continuum than a sharp dividing line
Our framework can be summarized using the schema in Table \ref{tab:schema}. We additionally note that we have limited our scope to only algorithms that work during training time, rather than those that augment LLM capabilities at inference time.

\subsubsection{Compute-Equivalent Gain}
The concept of compute-equivalent gain (CEG) was introduced by \citet{davidsonAICapabilitiesCan2023} as the measurement of how much additional training compute would be needed to improve benchmark performance by as much as the post-training enhancement. Concretely, we first estimate compute cost $C$ (in terms of floating point operations, FLOPs) as
\begin{equation}
\label{eq1}
    C \propto \textrm{Active Parameters Per Step} \times \textrm{Training Steps},
\end{equation}
Given the compute cost of a baseline model ($C_b$) and of a equally performant but more efficient model ($C_e$), we calculate the CEG as
\begin{equation}
\label{eq2}
    \textrm{CEG} = \frac{C_b}{C_e}.
\end{equation}
We note that our estimates using Equation \ref{eq1} and Equation \ref{eq2} do not holistically determine CEG, as they are unable to account for discrepancies in training data between models, and detailed training information for models is often not made public. Despite this, the rough estimates given by these equations fall in line with our experimental results and fit will within our framework.

\subsection{DeepSeek-V3: A First Case Study}
\label{sec:deepseek_case_study}
As a preliminary case study, we look at DeepSeek-V3, a model that notably required only 2.788 million H800 GPU hours for its full training \cite{deepseek-aiDeepSeekV3TechnicalReport2024}. In comparison, a similar state-of-the-art model, LLaMa 3.1-405B-Instruct, achieved comparable or worse performance with 30.84 million GPU hours \cite{schmidLlama31405B2025} on superior H100 GPUs \cite{IntroducingLlama31}. DeepSeek-V3's training time alone suggests that if compute advances were to suddenly stop, algorithmic improvements in model training could still increase capabilities. Perhaps more pertinently, DeepSeek-V3 also demonstrates that export controls cannot completely prevent the targeted countries from developing their own near-frontier AI models \cite{amodei2025deepseek, gupta_whackachip_2024a}. Of course, some have cast doubt on whether DeepSeek's training numbers are reliable. Ultimately, however, even DeepSeek's CEO stated that ``Money has never been the problem for us; bans on shipments of advanced chips are the problem'' \cite{schneiderDeepseekQuietGiant2024}. Either way, the algorithmic advancements used in DeepSeek-V3---including multi-headed latent attention, mixture-of-experts architecture, and mixed-precision training \cite{deepseek-aiDeepSeekV3TechnicalReport2024, daiDeepSeekMoEUltimateExpert2024, wangAuxiliaryLossFreeLoadBalancing2024}---no doubt contributed to its superior performance. DeepSeek-V3's performance suggests that even the most effective hardware controls cannot completely halt the improvement of LLMs in areas export controls affect.

\section{Classifying Algorithmic Advancements}

\subsection{Compute-Dependent Algorithms}
\subsubsection{Transformer}\label{sec:transformer}
The single most influential algorithmic advancement in the last decade is undoubtedly the transformer architecture \cite{vaswaniAttention}. In fact, some estimates \cite{hoAlgorithmicProgressLanguage2024} have suggested that the transformer architecture itself accounts for nearly 20\% of language modeling improvements since 2015.

The primary mechanism in the transformer is a multi-head self-attention layer. Self-attention allows the transformer model to draw global dependencies between input and output by tracking how each input token affects each of the others (or all previous tokens in causal self-attention) in a constant number of sequential operations.
As a result, the time and space complexity of self-attention scale as $O(n^2)$ where $n$ is the number of tokens. Reducing or circumventing this quadratic scaling has therefore been a major focal point of many other algorithms we analyze.
Previous state-of-the-art models, such as recurrent neural networks, required $O(n)$ \textit{sequential} operations to process a length-$n$ input, whereas the transformer architecture can process all tokens simultaneously, enabling greater throughput during training. 

Therefore, the key benefit of the transformer is its inherent parallelizability, stemming from self-attention's ability to process all token relationships with a fixed number of sequential operations. This makes the transformer highly parallelizable, although with a rather larger amount of overhead due to the multitude of self-attention layers in the model, each incurring the $O(n^2)$ complexity cost. This overhead is impractical at low levels of compute, explaining why previous research has found that other architectures, like long-short term memory (LSTM), are more efficient than transformers at smaller scales \cite{droppoScalingLawsAcoustic2021}. Conversely, the transformer's performance improves as the number of parameters and amount of data increases, as it can more efficiently utilize available compute. These facts together show that the transformer is compute-dependent: its advantages emerge in larger models trained with high enough compute.

We estimate the CEG of the transformer directly from \citet{vaswaniAttention}, which showed that the transformer performed as well as or better than all other state-of-the-art models on English-to-French and English-to-German translation tasks. The most performant English-to-French non-transformer model required $7.7 \times 10^{19}$ FLOPs to train, compared to the base transformer's $3.3 \times 10^{18}$, indicating that the transformer offers at least a $20 \times$ CEG. Further, the big transformer with $2.3 \times 10^{19}$ FLOPs achieved similar performance to the ConvS2S Ensemble at $1.2 \times 10^{21}$ FLOPs. This would put the transformer's CEG upwards of $50\times$, which is not unreasonable given previous findings about the transformer's influence relative to other algorithmic improvements \cite{hoAlgorithmicProgressLanguage2024}. Thus, we estimate the transformer's CEG to be between $20$ and $50\times$.

\subsubsection{Sparse Attention}
As previously discussed, one of the main drawbacks of the traditional self-attention mechanism is its quadratic complexity with respect to sequence length caused by each token interacting with every other token. While this issue is not of major concern with shorter sequences, it becomes a very pertinent issue once processing long sequences becomes the focus. To combat this problem, sparse attention \cite{childGeneratingLongSequences2019} was introduced to limit the size of the attention window, thereby reducing the mechanism's complexity. 

More specifically, whereas traditional (causal/masked) self-attention would ensure that all previous tokens attend to the current token, sparse attention selects only a subset of the previous tokens. \citet{childGeneratingLongSequences2019} chooses this subset such that it scales with $\sqrt{n}$, while still employing strategies (such as multi-hop attention paths) to ensure information can propagate across the full sequence. This gives sparse attention an improved complexity of $O(n\sqrt{n})$ over full self-attention's $O(n^2)$.

We classify sparse attention as compute-dependent by considering where its most transformative benefits are realized, particularly its ability to overcome dense attention's $O(n^2)$ limitations for exceptionally long sequences. The optimal application of sparsity involves a trade-off between efficiency and information flow, as highlighted by \citet{nawrotSparseFrontierSparse2025}, which found that for shorter sequences ($\sim32$k tokens), increasing attention density improves performance more efficiently than increasing model size with sparser attention. This suggests that the most aggressive sparse patterns, while essential for ultra-long sequences, impose constraints that are best managed when the demand for extreme context length is paramount. Such demanding scenarios typically require substantial compute for both processing and leveraging such vast input. Therefore, sparse attention's defining role in enabling ultra-long context capabilities at higher levels of compute aligns it with compute-dependent advancements.


When trained on the EnWiki8 dataset, the sparse transformer achieved $0.99$ bits per byte, matching that of the state-of-the-art model trained with more than double the number of parameters at half the time-per-iteration (i.e., sparse model at 0.55 units vs. SOTA at 1.33 units) \cite{childGeneratingLongSequences2019}. This gives a CEG estimate of $\frac{2 \cdot 1.33}{0.55} \approx 4.8\times$. More recently, \citet{yuanNativeSparseAttention2025} introduced a natively-trainable sparse attention mechanism with a number of additional optimizations, which achieved a $9.0\times$ speedup on the forward pass and $6.0\times$ speedup on the backward pass, for a CEG of $\approx 7\times$ (floor of the geometric mean due to greater compute required for the backward pass). We thus estimate sparse attention's CEG as $4.8$--$7\times$.

\subsubsection{Mixture of Experts}

The Mixture of Experts (MoE) architecture significantly increases a model's effective parameter count, and thus potential capabilities, without a proportional surge in computational cost (FLOPs) per processed token. This is achieved by employing a collection of specialized ``expert" sub-networks, where a routing mechanism dynamically activates only a small subset (often one or two) for each input token during a forward pass \cite{shazeerOutrageouslyLargeNeural2017, fedusSwitchTransformersScaling2022, daiDeepSeekMoEUltimateExpert2024}. While foundational ideas trace back earlier \cite{jacobsAdaptiveMixturesLocal1991}, MoE's successful deployment in modern LLMs such as DeepSeek-V3 depended on algorithmic innovations (e.g., for load balancing, although their necessity is debated \cite{wangAuxiliaryLossFreeLoadBalancing2024, daiDeepSeekMoEUltimateExpert2024}) and, crucially, on techniques enabling efficient parallelization of numerous experts across distributed hardware \cite{fedusSwitchTransformersScaling2022, lepikhinGShardScalingGiant2020}. Typically integrated into Transformers by replacing feed-forward network layers, MoE has demonstrated substantial performance advantages. For instance, \citet{fedusSwitchTransformersScaling2022} showed that a 7B parameter Switch Transformer, holding FLOPs per token constant, significantly outperformed a 200M parameter dense model. More recently, DeepSeekMoE's 16B-parameter model achieved performance comparable to the dense LLaMA2 7B model while using only 40\% of the compute \cite{daiDeepSeekMoEUltimateExpert2024}, underscoring MoE's parameter efficiency through specialization.

We classify MoE architectures as a compute-dependent advancement. This is primarily because its most significant benefits and highest CEG are realized in very large-scale models (e.g., trained with over $10^{23}$ FLOPs), where the massively increased parameter count is effectively leveraged by distributing and parallelizing experts across extensive hardware. While MoE can also be implemented at smaller scales, the overhead of routing, difficulty of effective expert specialization with less capacity, and engineering complexity for distributing experts may offer worse returns compared to well-optimized dense models at smaller scales. 

As previously discussed, DeepSeek-V3 exhibited an estimated CEG of approximately $11\times$ over LLaMa 3.1-405B given their similar performance and difference in training times (although there are a number of confounding factors in this comparison, including GPU differences and other algorithmic improvements) \cite{deepseek-aiDeepSeekV3TechnicalReport2024, schmidLlama31405B2025, IntroducingLlama31}. As a more reliable estimate, a FLOP-matched Switch Transformer also showed a $\approx 7\times$ training speedup (which we take as our CEG estimate) over a base dense model \cite{fedusSwitchTransformersScaling2022}. This strong performance at scale leads to MoE models being primarily deployed as alternatives to the largest dense transformers and suggest the compute-dependence of the algorithmic improvement.

We note, however, that MoE architectures present a unique challenge for our CEG framework (that also points to their compute-dependence). While active FLOPs per token are kept low, the total parameter count is vast. This necessitates significant memory, storage, and high-bandwidth communication infrastructure—hallmarks of high-compute environments—to manage and efficiently operate these distributed models. This reliance on substantial infrastructure for the full model, beyond just active FLOPs, reinforces their compute-dependent classification. It is important to note, however, that MoE principles showed utility even in earlier, smaller-scale models \cite{shazeerOutrageouslyLargeNeural2017} before massive transformer scaling, indicating the core algorithmic concept offers benefits more broadly, even if its most transformative impact and current deployments are concentrated at the higher end of the compute spectrum.

\subsubsection{Multi-Query Attention}
To reduce memory usage in the attention mechanism, multi-query attention (MQA) modifies standard multi-head attention (where each head has its own query, key, and value projections) by having all query projections share the same key and value projections \cite{shazeerFastTransformerDecoding2019}. This reduces the KV cache size by a factor of the number of attention heads while maintaining multiple query projections, resulting in comparable model quality to standard multi-head attention. For example, on the WMT14 English-to-German translation task, MQA achieved a BLEU score of 28.5, nearly matching the baseline score of 28.4, with only a slight increase in perplexity (1.439 vs. 1.424) \cite{shazeerFastTransformerDecoding2019}.

Architecturally, MQA reduces the number of key and value projection parameters compared to MHA, leading to a slight decrease in FLOPs per training step. However, the original work by \citet{shazeerFastTransformerDecoding2019} reported that overall training step-times for MQA and Multi-Head Attention (MHA) were comparable on the models tested (13.0 $\mu$s vs. 13.2 $\mu$s per token). This suggests that MQA's direct CEG is roughly neutral (i.e., $\approx 1.0$) at the scale and context of its introduction.

The primary impact of MQA, however, lies in its substantial improvements to inference efficiency by dramatically reducing the size of the KV cache and associated memory bandwidth requirements. \citet{shazeerFastTransformerDecoding2019} demonstrated significant inference speedups, particularly for autoregressive decoding where decoder inference time was reduced from 46 $\mu$s to an impressive 3.8 $\mu$s per token (an approximate $12.1\times$ speedup), with more modest gains for encoder tasks ($1.13\times$). This substantial inference efficiency becomes even more critical and its benefits more pronounced as models become larger and deployable contexts lengthen, making KV cache management a primary operational bottleneck. This advantage has made MQA a crucial optimization in modern LLMs including Falcon, PaLM, and StarCoder \cite{malarticFalcon211BTechnicalReport2024,chowdheryPaLMScalingLanguage2022,liStarCoderMaySource2023}, despite its initially slow adoption. Further, \citet{shazeerFastTransformerDecoding2019} found a increase in perplexity in MQA compared to standard multi-head attention. Therefore, the trade-offs involved in sharing K/V projections, while beneficial for inference at larger scales, may not be optimal for model quality at smaller scales where memory constraints are less pressing. Thus, with MQA's most significant advantage (inference efficiency) escalating in importance as model scale and computational demands increase, we classify it as a compute-dependent advancement.

\subsection{Compute-Independent Algorithms}

\subsubsection{Rotary Positional Embeddings}
The self-attention mechanism models interactions between tokens in a sequence without regard to their order, a property known as permutational equivariance. As a result, the algorithm does not ``know" whether a token appears at the beginning or end of a sentence. However, since natural language is inherently sequential, this issue is typically addressed using positional embeddings, which encode information about a token's location within a sequence.

In the original transformer architecture, positional embedding was implemented using periodic functions (sines and cosines) with geometrically spaced frequencies up to a fixed maximum sequence length \cite{vaswaniAttention}. While simple and efficient, this approach has two major limitations that led to the development of rotary positional embedding (RoPE). First, absolute encoding represents a token's position in isolation rather than encoding its relative distance to other tokens, making it less effective for capturing relationships in long sequences. Second, the use of a fixed maximum sequence length imposes constraints on how much text the model can process effectively, leading to degradation in quality when modeling long-range dependencies.

To circumvent these issues, RoPE \cite{suRoFormerEnhancedTransformer2023} was introduced as a way to model relative sequence positions using rotation matrices instead of traditional additive embeddings. RoPE applies rotational transformations directly to the query and key embedding vectors based on their absolute positions within the sequence. Consequently, when the attention between a rotated query (from position $m$) and a rotated key (from position $n$) is computed, the resulting score inherently becomes a function of their relative position ($m - n$) on top of their original content. The mathematical properties of these rotations allow the encoded inter-token dependency to diminish naturally with increasing sequence distance, aligning well with typical modeling objectives for sequential data. Furthermore, RoPE adds negligible computational overhead compared to absolute positional encoding schemes, as it primarily involves efficient element-wise operations and does not introduce new learnable parameters.

RoPE presents a clear example of a compute-independent algorithmic advance, as the improvements can be observed in all scales of transformer architectures. RoPE has been incorporated in many language models after its original publication, and has been further extended to handle even longer sequence lengths (e.g. LongRoPE \cite{dingLongRoPEExtendingLLM2024} and other positional embeddings such as AliBi \cite{pressTrainShortTest2022}). RoPE is essentially a step improvement in context length; it achieves the same performance for a much larger number of tokens, provides a greater ability to summarize long texts.

We estimate RoPE's CEG directly from benchmarks in the original paper. The RoFormer paper trained BERT (110M base parameters \cite{devlinBERTPretrainingDeep2019}) and RoFormer (65M base parameters) over 100K training steps. Both models exhibited comparable performance on downstream GLUE tasks. Consequently, we estimate the computational cost for each model as 
$C_{\textrm{BERT}} = (110 \cdot 10^6) \cdot (1 \cdot 10^5) = 1.1 \cdot 10^{13}$ and $C_{\textrm{RoFormer}} = (65 \cdot 10^6) \cdot (1 \cdot 10^5) = 6.5 \cdot 10^{12}$. Thus, RoPE's estimated CEG is $\frac{1.1 \cdot 10^{13}}{6.5 \cdot 10^{12}} \approx 1.7 \times$.

\subsubsection{FlashAttention}
As previously discussed, the self-attention mechanism scales quadratically with the length of the sequence. Such quadratic scaling can easily become prohibitive when handling long sequences required for complex language processing tasks.
FlashAttention \cite{daoFlashAttentionFastMemoryEfficient2022} is an algorithm aimed to improve the efficiency of the attention mechanism by taking advantage of the physical memory layout on the GPUs used for training. In a modern GPU, the computing speeds are significantly faster than the speeds at which data is transferred, meaning that only a fraction of time is actively spent doing the attention calculation during a typical self-attention calculation \cite{jia_dissecting_2018}. In FlashAttention, this difference in speed can be leveraged to compute the attention matrix in chunks, rather than transferring it all at once to and from the slow (but large) GPU memory. The result of these hardware-aware optimizations is approximately a 2--4$\times$ speedup compared to the default PyTorch implementation of attention and enables sequence lengths around 4 times longer than the original. Later iterations of FlashAttention further develop hardware-based optimizations to increase the speed and long-sequence capabilities \cite{daoFlashAttention2FasterAttention2023, shahFlashAttention3FastAccurate2024}. FlashAttention is now a primary algorithm used in state-of-the-art language modeling due to its performance efficiency and their widespread use of attention mechanisms.

FlashAttention shows time and memory improvements at all sequence lengths tested (see Fig. 3 in \citet{daoFlashAttention2FasterAttention2023}). This is also generally true for later generations, suggesting that the algorithmic innovation was largely compute-independent. In this specific case, it appears that algorithmic optimization can significantly reduce compute requirements without sacrificing model accuracy or capabilities. It must be noted, however, that FlashAttention takes special advantage of the hardware layout found on modern GPUs (specifically the multiple levels of memory between the fast SRAM and slow high bandwidth memory), making it highly dependent on hardware architecture and the ubiquity of GPUs for AI training. Despite this hardware dependence, the widespread adoption and utility of FlashAttention across the spectrum of model sizes commonly trained within this GPU ecosystem lead us to classify it as primarily compute-independent, as its core advantages are realized across a range of compute scales.

To estimate FlashAttention's CEG, we note that FlashAttention primarily yields improvements through hardware-aware optimizations that increase wall-clock speed, rather than reducing the number of active parameters or decreasing the number of training steps required for convergence. Therefore, under a strict application of our CEG definition, FlashAttention's CEG would be $1\times$. However, its practical impact on compute efficiency is substantial. The previously mentioned 2--4$\times$ speedup in training wall-clock time compared to standard PyTorch attention implementations reflects a significant real-world compute saving. While this speedup metric does not directly map to our CEG equation, it represents the crucial efficiency gain that drives FlashAttention's widespread use. Therefore, while the formal CEG is $1\times$, we report this 2--4$\times$ speedup as the key metric representing its substantial real-world efficiency.

\subsubsection{Layer Normalization}
Prior to layer normalization \cite{baLayerNormalization2016}, batch normalization was the standard for training neural networks. Batch normalization consists of computing the mean and variance of each feature within a mini-batch, then using these statistics to normalize each input's features. For natural-language processing (NLP) applications, however, batch normalization can struggle due to variable mini-batch sizes, meaning that the computed statistics may be inaccurate estimators. Despite this drawback, batch normalization provides a significant reduction in the training time of neural networks that utilize the technique.

Layer normalization provides training speedups similar to batch normalization in a more reliable manner for NLP tasks. Rather than using the mean and variance of each minibatch to normalize, it uses the mean and variance of the features within a single input \cite{baLayerNormalization2016}. Layer normalization is therefore not prone to being influenced by the batch size. Additionally, layer normalization empirically yields training speedups, especially when given long sequences or small batch sizes. 

Given that layer normalization demonstrated speedups on the much smaller neural networks used in 2016 and is still often used in models today (or is swapped out for similar normalization techniques, such as RMSNorm \cite{zhang_root_2019}), this algorithmic advancement falls into the compute-independent category. This is further supported by layer normalization's strong theoretical motivation \cite{baLayerNormalization2016}; no advances in compute were needed to demonstrate the improvement this algorithm provides.

We estimate the CEG offered by layer normalization directly from \citet{baLayerNormalization2016}, which demonstrated that an LSTM with layer normalization could converge using approximately 60\% of the training steps required by the base model without layer normalization. Thus, layer normalization's CEG is $\frac{1.0}{0.6} \approx 1.67\times$.

\subsection{Results}

We summarize our findings in Table \ref{findings}. Our findings indicate that numerous algorithmic advancements have provided non-trivial increases in CEG, implying that if compute power were to be frozen tomorrow, LLM progress could still continue from pretraining algorithmic advancements alone. Therefore, these results highlight the potential limitations of AI governance strategies focused solely on restricting compute access, as compute-independent advancements could still provide meaningful improvements to compute-restricted models.
We additionally observe that compute-dependent advances tend to provide greater compute-equivalent gain than compute-independent advances, suggesting that there is an interaction between compute increases and furthering algorithmic progress.

\begin{table*}[t]
\centering
\caption{Summary of estimated CEG for reviewed algorithmic advancements. ``Speedup'' indicates improved wall-clock performance from memory efficiency, not decreases in FLOPs.}
\begin{tabular}{|l|l|l|}
\hline
\textbf{Advancement} & \textbf{CEG} & \textbf{Compute Dependent/Independent} \\ \hline
Transformer & $20$--$50\times$ & Dependent \\ \hline
Sparse Attention & $4.8$--$7\times$ & Dependent \\ \hline
MoE & $7$--$11\times$ & Dependent \\ \hline
MQA & \begin{tabular}{@{}l@{}}$1\times$ \\ $1.13\times$ encoder speedup \\ $12.1\times$ decoder speedup\end{tabular} & Dependent \\ \hline
RoPE & $1.7\times$ & Independent \\ \hline
FlashAttention & \begin{tabular}{@{}l@{}}$1\times$ \\ $2$--$4\times$ practical speedup\end{tabular} & Independent \\ \hline
LayerNorm & $1.67\times$ & Independent \\ \hline
\end{tabular}
\label{findings}
\end{table*}

\section{Experiments}

To empirically verify our analysis, we conducted a series of experiments using two scales of small models to measure the effects of different algorithmic improvements across different compute scales. Due to monetary constraints, we were unable to run large-scale experiments, however, we found that even the smaller scale validated our framework.

We trained a scaled-down version of OpenAI's GPT-2 model \cite{radford_language_2019} based on the PyTorch implementation nanoGPT \cite{andrej_karpathy_2025}. 
Due to limited computing resources, we trained our model on the OpenWebText \cite{pile} dataset for a fixed number of iterations (50,000) rather than to convergence, which requires approximately 600,000 iterations for the full-size nanoGPT model. We implemented each algorithm using the native PyTorch libraries (LayerNorm, FlashAttention) or as closely to the original report as possible otherwise (sparse attention, MQA, and RoPE). Given the architectural demands and our model scales, mixture-of-experts was not implemented and is thus excluded from our direct experimental results.

We tested each algorithmic advancement on two scales: a compact model ($\sim$50M parameters) and a full model ($\sim$110M parameters). These separate sizes allow us to observe how CEG scales with increased computational budget. For each experiment, an algorithmic advancement was applied to a baseline model configuration, and its cross-entropy validation loss was measured after 50,000 training iterations. Within each model size category (compact or full), the underlying architectural parameters determining FLOPs per iteration were kept consistent across all algorithmic variations. This ensured that each 50,000-iteration run corresponded to a comparable computational budgets (total FLOPs executed) for that specific model size, allowing for simple comparison of validation loss achieved for the same compute.

To quantify performance differences, we estimated the Compute-Equivalent Gain (CEG). Our primary method defines a target validation loss, $L_{\mathrm{target}}$, as the loss achieved by the baseline model at its 50,000th iteration. Let $S_{\mathrm{base}}$ represent these 50,000 steps (which are directly proportional to FLOPs, as FLOPs/step is constant for a given model size). For any other algorithm under test, we identified the number of steps, $S_{\mathrm{alg}}$, it required to first achieve a validation loss less than or equal to $L_{\mathrm{target}}$. The Primary CEG was then calculated as $\frac{S_{\mathrm{base}}}{S_{\mathrm{alg}}}$.  Consequently, a CEG $> 1.0$ indicates the algorithm reached $L_{\mathrm{target}}$ in fewer steps (i.e., with less compute) than the baseline, signifying improved efficiency. Conversely, a CEG $< 1.0$ indicates the algorithm was less efficient in reaching $L_{\mathrm{target}}$.

If an algorithm did not reach $L_{\mathrm{target}}$ within its 50,000 training steps (its final loss was worse than $L_{\mathrm{target}}$), its primary CEG relative to $L_{\mathrm{target}}$ could not be meaningfully computed as above. In such cases, we report an auxiliary CEG. This auxiliary CEG is calculated by using the algorithm's own minimum achieved validation loss ($L_{\mathrm{alg}}$, achieved at $S'_{\mathrm{alg}}$ steps) as an alternative target. We then determine the steps the baseline model required to achieve this $L_{\mathrm{alg}}$ (denoted $S'_{\mathrm{base}}$) and compute the CEG as $\frac{S'_{\mathrm{base}}}{S'_{\mathrm{alg}}}$. This auxiliary CEG provides a measure of relative efficiency even when the primary performance target is not met by the algorithm under test.

\subsection{Experimental Results}
\label{sec:results}

\begin{figure*}[htb]
    \centering
    \caption{Training and validation loss curves during GPT training experiments across different algorithmic enhancements. The cross-entropy loss was reported every 500 iterations. FA: FlashAttention; LN: LayerNorm, SA: Sparse Attention.}
    \includegraphics{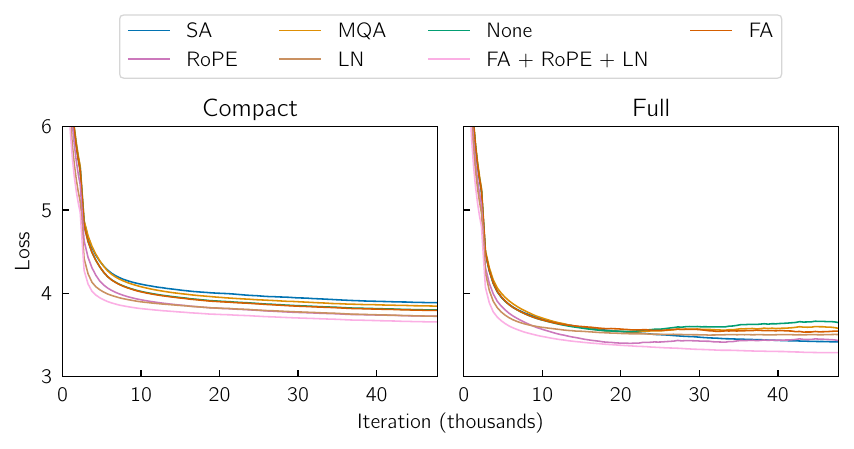}
    \label{loss}
\end{figure*}

\begin{table*}[ht]
\centering
\caption{Experimental Results}
\begin{tabular}{|c|c|c|c|c|}
\hline
\textbf{Algorithm} & \textbf{Min. Val. Loss} & \textbf{CEG} & \textbf{Min. Val. Loss} & \textbf{CEG} \\
\hline
& \multicolumn{2}{c|}{\textit{Compact Model}} & \multicolumn{2}{c|}{\textit{Full Model}}\\
\hline
Multi-Query Attention
& 3.846 & 0.673 
& 3.535 & 0.931 \\
Sparse Attention
& 3.886 & 0.515 
& 3.418 & 0.964 \\
None 
& 3.799 & 1.000 
& 3.542 & 1.000 \\
FlashAttention 
& 3.794 & 1.063 
& 3.532 & 1.000 \\
LayerNorm 
& 3.725 & 1.836 
& 3.498 & 1.421 \\
RoPE
& 3.725 & 1.870 
& 3.398 & 1.350 \\
LN + RoPE + FA
& 3.656 & 3.483 
& 3.287 & 1.800 \\
\hline
\end{tabular}
\label{algo_comparison}
\end{table*}

\paragraph{Compute-Independent Algorithms}
Our experiments confirmed that algorithms classified as compute-independent demonstrated improved training efficiency 
(CEG $>$ 1.0) even in the compact model. For the compact model, layer normalization and RoPE yielded CEG values greater than 1.0 ($1.836\times$ and $1.870\times$, respectively), indicating a reduction in the FLOPs required to reach a given loss. These benefits were also observed in the full model (albeit to a slightly lesser extent). 
FlashAttention demonstrated a CEG of $1.063\times$, which falls in line with our previous estimated CEG of $\approx 1\times$. Notably, however, FlashAttention did reduce training time in both models, again validating our previous analysis. Namely, wall-clock training time was reduced from 106 to 83 minutes for the compact model and from 484 to 320 minutes for the full model, giving FlashAttention a speedup range of $1.27$--$1.51\times$.

\paragraph{Compute-Dependent Advancements}
The compute-dependent advancements we tested exhibited a distinct scaling pattern, aligning with our framework's expectations. 
In the compact model, these algorithms (MQA and sparse attention) resulted in a CEG less than $1.0$, with MQA exhibiting a CEG of $0.673\times$ and sparse attention a CEG of $0.515\times$. Thus, at this scale they were detrimental to training efficiency, requiring more FLOPs than the baseline to achieve the same loss. However, when applied to the full model, their CEG improved nearly to 1.0 (MQA CEG: $0.931\times$, sparse attention CEG: $0.964\times$), meaning they performed comparably to the baseline, no longer degrading FLOP efficiency. This behavior---poor or neutral performance at smaller scales improving towards neutrality or potential benefit with increased compute---is characteristic of compute-dependent advancements as defined in our framework.

Specifically for multi-query attention, while we had previously discussed its inference-time benefits (primarily when decoding), our analysis of its training-time FLOP efficiency still categorizes it as compute-dependent. Its scaling behavior in our experiments (degrading FLOP efficiency at small model sizes, becoming neutral at larger model sizes for training) supports this classification, irrespective of its known advantages in deployment.

\section{Discussion}

Our case-by-case analysis of selected algorithms demonstrates that compute-independent algorithmic advancements do exist, but the most impactful algorithmic advancements tend to be compute-dependent (Table~\ref{findings}). Moreover, these findings are generally confirmed by our empirical tests: the compute-dependent advances at best did nothing and at worse slightly degraded our small-scale model's performance, whereas the compute-independent advances improved performance at all scales (Table~\ref{algo_comparison}). 

Our distinction between compute-dependent and compute-independent algorithmic advancements offers a valuable lens through which to consider long-term AI trajectories and potential shifts in research priorities.
While historical progress in frontier models has heavily relied on compute scaling \cite{epoch2024trainingcompute, sevilla2022computetrends}, understanding the nature of algorithmic innovation becomes even more critical if such scaling faces future limitations.
Should data or hardware availability plateau, future research in language modeling would likely intensify its focus on compute-independent advances that drive efficiency. This could include a greater emphasis on architectural innovations or training methodologies that yield better performance per FLOP. Simultaneously, there would be increased value in discovering ways to make traditionally compute-dependent breakthroughs more algorithmically efficient, lessening their reliance on increasing scale.

We further note that the exact cause for some algorithmic advancements being compute-dependent while others are compute-independent is unclear, however there exist some commonalities between the algorithms within each group.
Among the compute-dependent advancements we analyzed, we observe that each advancement directly affects the attention mechanism. For example, the transformer architecture introduced the attention mechanism, and in MoE, the attention mechanism is ``split'' across multiple experts. Conversely, compute-independent advances tend to operate independently of the attention mechanism. (Even though its name suggests otherwise, FlashAttention changes how the algorithm is implemented in the hardware and does not change the mechanism itself.) Our findings therefore suggest that the attention mechanism is the primary compute consumer (which falls in line with its $O(n^2)$ scaling), and that improvements to the mechanism consequently yield the greatest compute-equivalent gains. This in turn suggests that improving the attention mechanism should be a focus of further research. However, given our study's narrow focus on language modeling, this hypothesis might simply be a spurious correlation, influenced by selection bias and the frequent innovations in attention mechanisms characteristic of this field

Finally, we note that our compute-dependent vs. compute-independent framework can serve as a useful framework for directing future AI research. If, before testing, researchers hypothesize than an algorithmic advancement is compute-independent, initial experiments could be run at smaller scales to confirm the advancement's efficacy. On the other hand, if the advancement is hypothesized to be compute-dependent, researchers should recognize that its defining benefits might only become evident at larger scales. This could guide their experimental trajectory, emphasizing the need for eventual larger-scale validation to test the advancement's true impact, rather than prematurely dismissing it based on solely small-scale outcomes. We therefore hope that going forward, this framework can be used to speed up the empirical validation step of the research process.

\subsection{Implications for Policymakers and Researchers}

The existence of compute-independent advancements implies that even at lower compute levels, models can be improved through algorithmic innovation alone. Even if the total available compute were to suddenly freeze, researchers could still improve models via compute-independent advances.
Therefore, hardware controls, while clearly impactful, are no silver bullet. Even if the strictest controls were enacted and enforced, this would not suffice to stifle language modeling progress \cite{gupta_whackachip_2024a}. The emergence of powerful models like DeepSeek-V3, developed under certain hardware constraints, highlights that significant algorithmic advancements can still enable near-frontier capabilities, challenging the notion that hardware access alone dictates the pace of progress at the cutting edge. (We reiterate here the DeepSeek's CEO's comments on the availability of compute.)

At the same time, however, because the biggest capability gains (for us, CEGs) appear to come from compute-dependent advancements, hardware controls are not entirely futile. By restricting the large-scale compute necessary to validate and refine the substantial CEG offered by these compute-dependent breakthroughs, such controls will likely serve to temper the pace of cutting-edge algorithmic development in affected areas.

To the extent that algorithmic advancements that improve the performance of small models also improve the performance of larger models, we also expect research and development to be easier. If results from experiments on smaller models translate to larger models, it will take less compute, researcher hours, and time to iterate, and it will be easier to run many experiments in parallel. If the bar for humans to run experiments is lower, it would also lower the threshold for effective automation of these experiments, and we thus expect automated search processes to be especially potent in identifying these compute-independent advances.

Though the lower experimental bar allows less-resourced researchers to identify and verify the effectiveness of compute-independent advances (since their benefits are evident at small scales), this also means that better-resourced researchers can search for a larger number of such validated advances through more extensive experimentation.
In this way, organizations which already have large amounts of compute may counterintuitively be better positioned to discover new compute-independent advances. They have the resources to search more comprehensively across the range of possible compute-independent advances, and can likely automate their search more easily than competitors with less computation power. Indeed, this approach sounds remarkably similar to the one described by Google's Chief Scientist Jeff Dean during his February 2025 appearance on the Dwarkesh Podcast:

\begin{quote}
    I think one thing people should be aware of is that the improvements from generation to generation of these models often are partially driven by hardware and larger scale, but equally and perhaps even more so driven by major algorithmic improvements and major changes in the model architecture, the training data mix, and so on, that really makes the model better per FLOP that is applied to the model. \dots Then I think if we have automated exploration of ideas, we'll be able to vet a lot more ideas and bring them into the actual production training for next generations of these models.
    
    That's going to be really helpful because that's sort of what we're currently doing with a lot of brilliant machine learning researchers: looking at lots of ideas, winnowing ones that seem to work well at small scale, seeing if they work well at medium scale, bringing them into larger scale experiments, and then settling on adding a whole bunch of new and interesting things to the final model recipe. If we can do that 100 times faster through those machine learning researchers just gently steering a more automated search process, rather than hand-babysitting lots of experiments themselves, that's going to be really, really good. \cite{patel2024dean}
\end{quote}

Another key consequence of our findings for policymakers who are interested in the most-capable models is that it will be more difficult to define the frontier with pre-training compute alone. Record-breaking quantities of pre-training compute remain sufficient for frontier capabilities, but are less necessary. In addition to model distillation (not discussed here) which let smaller models achieve performance comparable to larger models by training on their outputs, algorithmic improvements could allow models to increase performance without crossing compute thresholds that trigger governance requirements. 

Addressing this gap presents a profound challenge for which current governance tools appear inadequate. Attempting to monitor or shape algorithmic research is one option, however this is not currently widespread practice, largely due to significant political and practical challenges (including concerns about innovation and oversight capacity). 
Evaluations may be feasible for particular narrow capabilities \cite{karnofsky2024tripwire}, but the application of politically-accepted regulatory tripwires based on assessments of general capabilities is not yet established and faces substantial hurdles in terms of measurement and consensus.

\subsection{Limitations and Future Directions}
There are several limitations to our analysis. First, we do not account for the role of data improvements LLM advancements, which may be particularly relevant for models trained on other LLM outputs. This contrasts with prior work, such as Epoch AI's dataset-equivalent gain estimates \cite{hoAlgorithmicProgressLanguage2024}. Additionally, we do not consider inference-time improvements or reinforcement learning techniques, such as chain-of-thought prompting \cite{wei_chainofthought_2023}, reinforcement learning from human feedback \cite{ouyang_training_2022}, or DeepSeek's group-relative policy optimization methods \cite{deepseek-ai_deepseekr1_2025}, which may further impact LLM capabilities.

Furthermore, our own empirical investigations faced specific constraints. Due to limited computational resources, experiments were conducted on models of a relatively modest scale (i.e., 50M to 110M parameters), which is substantially smaller than current frontier models. For the same reason, these experimental models were not always trained to full convergence. These factors could influence the precise CEG values reported and the observed performance characteristics of some algorithms, particularly for compute-dependent advances whose benefits often become more pronounced at larger scales or with more extensive training.

Another key limitation is our assumption that algorithmic advancements act independently, when in reality, they often are applied together. Our experiment did not test every possible algorithmic combination, though the superior performance of LayerNorm + RoPE + FlashAttention suggests the existence of cooperative effects between algorithms. Similarly, our estimates rely on evaluating algorithms in isolation from state-of-the-art models, whereas modern models integrate multiple advancements (e.g., LayerNorm, RoPE, and MQA, often with modifications like RMSNorm or LongRoPE) \cite{zhao_survey_2024}. Consequently, our estimates may underestimate the true impact of algorithmic progress.

To address this study's limitations, future research should aim to more comprehensively assess algorithmic progress. Key directions may include developing methods to quantify the impact of data improvements (especially model-generated data) in conjunction with algorithmic changes; integrating the effects of inference-time optimizations and reinforcement learning techniques into analyses; and more holistically evaluating algorithmic contributions by studying their synergies. Furthermore, investigation into how algorithmic advancements affect model distillation and continued expansion of the analytical framework to incorporate new categories will be crucial to more accurately map the trajectory of LLM advancements.

\section{Conclusion}
In this paper, we introduced a novel framework for classifying algorithmic advancements as either compute-dependent or compute-independent, providing a clearer understanding of how LLM capabilities can progress even under hardware constraints. Our empirical validation confirmed that compute-independent innovations like Layer Normalization, RoPE, and FlashAttention yield meaningful performance gains (up to $3.5\times$ compute-equivalent gain) even in resource-constrained settings. On the other hand, the most impactful advancements remain compute-dependent and often attention-focused, suggesting that export controls may slow, but cannot fully prevent, AI progress. Our investigation focused on algorithmic improvements to pretraining, and future work should address advancements in other places, for example, by estimating the compute-equivalent gain from chain-of-thought prompting. Such estimates might give us a more holistic view of the role non-architectural factors play in the increasing performance of frontier models.

\section{Acknowledgments}
We wish to acknowledge the support of the University of Chicago's Existential Risk Laboratory.

\bibliography{aaai25}



\end{document}